\documentclass[lettersize,journal]{IEEEtran}

\ifCLASSOPTIONcompsoc
  \usepackage[nocompress]{cite}
\else
  \usepackage{cite}
\fi
\ifCLASSINFOpdf
\else
\fi
\usepackage[numbers]{natbib}
\usepackage{hyperref}
\usepackage{enumitem}
\usepackage{amsmath,amssymb,amsfonts}
\usepackage{booktabs, multirow} 
\usepackage{soul}
\usepackage[table]{xcolor} 
\usepackage{changepage,threeparttable} 
\usepackage{caption}
\usepackage{subcaption}
\usepackage{adjustbox}
\usepackage{epstopdf}
\usepackage{amsmath}
\usepackage{algorithm}
\usepackage{algorithmic}
\usepackage{tikz}
\usepackage{pgfplots}
\usepackage{pgfplotstable}
\pgfplotsset{compat=1.7}
\usetikzlibrary{arrows.meta,
                chains,
                positioning}
\usepackage{amsmath}

\begin{document}
%
\title{QXAI: Explainable AI Framework for Quantitative Analysis in Patient
Monitoring Systems}
%
%
%
%

\author{Thanveer Shaik,  Xiaohui Tao, Haoran Xie, Lin Li, Juan D. Velásquez, and Niall Higgins
\thanks{Thanveer Shaik and Xiaohui Tao are with 
the School of Mathematics, Physics \& Computing, University of Southern Queensland, Toowoomba, Queensland, Australia (e-mail: Thanveer.Shaik@usq.edu.au, Xiaohui.Tao@usq.edu.au).}
\thanks{Haoran Xie is with the Department of Computing and Decision Sciences, Lingnan University, Tuen Mun, Hong Kong (e-mail: hrxie@ln.edu.hk)}
\thanks{Lin Li is with the School of Computer and Artificial Intelligence, Wuhan University of Technology, China (e-mail: cathylilin@whut.edu.cn)}
\thanks{Juan D. Velásquez is with the Industrial Engineering Department at University of Chile, Chile (e-mail: jvelasqu@dii.uchile.cl)}
\thanks{Niall Higgins is with Metro North Hospital and Health Service, Royal Brisbane and Women’s Hospital, and also  with School of Nursing, Queensland University of Technology, Brisbane, Australia (e-mail: Niall.Higgins@health.qld.gov.au).}
}

\IEEEtitleabstractindextext{%
\begin{abstract}
Artificial Intelligence techniques can be used to classify a patient's physical activities and predict vital signs for remote patient monitoring. Regression analysis based on non-linear models like deep learning models has limited explainability due to its black-box nature. This can require decision-makers to make blind leaps of faith based on  non-linear model results, especially in healthcare applications. In non-invasive monitoring, patient data from tracking sensors and their predisposing clinical attributes act as input features for predicting future vital signs. Explaining the contributions of various features to the overall output of the monitoring application is critical for a clinician's decision-making. In this study, an Explainable AI for Quantitative analysis (QXAI) framework is proposed with post-hoc model explainability and intrinsic explainability for regression and classification tasks in a supervised learning approach. This was achieved by utilizing the Shapley values concept and incorporating attention mechanisms in deep learning models. We adopted the artificial neural networks (ANN) and attention-based Bidirectional LSTM (BiLSTM) models for the prediction of heart rate and classification of physical activities based on sensor data. The deep learning models achieved state-of-the-art results in both prediction and classification tasks. Global explanation and local explanation were conducted on input data to understand the feature contribution of various patient data. The proposed QXAI framework was evaluated using PPG-DaLiA data to predict heart rate and mobile health (MHEALTH) data to classify physical activities based on sensor data. Monte Carlo approximation was applied to the framework to overcome the time complexity and high computation power requirements required for Shapley value calculations. 
\end{abstract}

\begin{IEEEkeywords}
Explainability, Shapley, Attention, Monte Carlo, Vital Signs, Physical Activities
\end{IEEEkeywords}}

\maketitle

\IEEEdisplaynontitleabstractindextext

%
\IEEEpeerreviewmaketitle

\section{Introduction}

In the realm of modern healthcare, the integration of cutting-edge technology, notably through remote monitoring systems, represents a pivotal advancement in patient care and the management of diseases. These systems play an essential role in the prompt detection and averting of grave health events, chiefly through their capacity to precisely monitor and scrutinize vital signs such as temperature, pulse, respiratory rate, and mean arterial pressure~\cite{malasinghe2019remote, asiimwe2020vital}. However, the scope of traditional monitoring systems is often constrained to displaying a patient's current health status, which limits their effectiveness in preemptively predicting and managing potential health complications.

The advent of Artificial Intelligence (AI) and deep learning heralds a new era in healthcare, transcending the boundaries of traditional methods by offering predictive insights that are indispensable for early and effective medical interventions~\cite{tao2022multi, prakash2020mapping}. Nevertheless, these advanced methodologies come with their own set of complexities, chief among them being the lack of transparency and comprehensibility in deep learning models. These models, often labeled as "black-box" models, pose a significant challenge in elucidating how input factors correlate with the predictive outcomes~\cite{muddamsetty2022visual, Adadi2020}. This issue is particularly critical in healthcare, where understanding the rationale behind AI-driven decisions is vital for their acceptance in clinical settings and for ensuring ethical applications of such technologies.

In response to these challenges, our research presents an innovative Explainable AI framework tailored for Quantitative data (QXAI), ingeniously amalgamating the Shapley values concept~\cite{lundberg2017unified} with an attention mechanism in the realm of deep learning models. Our approach is uniquely poised to demystify AI predictions on both granular (local) and aggregate (global) scales. It provides insightful revelations on how each individual feature contributes to specific input records and offers a comprehensive overview of feature contributions throughout the entire model. This dual-level explanation capability is adeptly employed in our framework for the purpose of predicting human vital signs and classifying physical activities, utilizing two advanced deep learning models: Artificial Neural Networks (ANN) and attention-based Bidirectional LSTM (BiLSTM). The empirical evidence from our study highlights the framework's proficiency in delivering detailed Shapley values and attention weights for each input feature, thereby clarifying their respective impacts on the outcomes of deep learning models.

Recognizing the computational demands in calculating Shapley values for extensive datasets, we have judiciously integrated the Monte Carlo method of approximation with random sampling. This strategic addition not only mitigates the computational complexities but also augments the practical utility of our framework across a spectrum of real-world applications.

Overall, our study represents a significant advancement in the field of explainable AI within healthcare. The key contributions of our research are:
\begin{itemize}
    \item Development of an innovative, adaptable Explainable AI framework (QXAI) for quantitative data analysis in healthcare. This framework uniquely combines attention layer mechanisms with Shapley values within deep learning models, setting a new standard in AI explainability.
    \item Comprehensive evaluation of the framework's explainability capabilities, focusing on the importance of features and providing both local and global explanations. This dual approach significantly enhances the understanding of AI models, offering insights into their cognitive and behavioral aspects.
    \item Adoption of the Monte Carlo method to address the computational challenges in calculating Shapley values, especially for large datasets. This method significantly reduces the computational overhead, making the framework more practical for real-world applications.
    \item Establishment of a new paradigm in patient monitoring systems for interpreting and explaining AI predictions related to vital signs and physical activity classification. This advancement is pivotal for clinical decision-making, offering a more nuanced and in-depth understanding of patient health dynamics.
\end{itemize}

The remainder of the article is organized as follows: Section~\ref{relatedwork} presents related works on explainability in healthcare applications. Section~\ref{problem} presents a formal definition of the research problem addressed. Section~\ref{methods} details the novel QXAI framework to explain prediction and classification problems proposed in this study. Experimental design, dataset description, data modelling, and traditional models are discussed in Section~\ref{exp-design}. In Section~\ref{results}, experimental results of the QXAI framework are discussed, along with its explainability and feature identification performance. Section~\ref{mcApprox} discusses the random sampling approximation using the Monte Carlo method. In Section~\ref{discussion}, we discuss implications, strengths, and limitations of the study. Finally, the paper concludes with Section~\ref{conclusion}.

\section{Related Work}\label{relatedwork}
In the realm of remote patient monitoring systems, the primary objective is to promptly identify high-risk patients, enabling clinicians to allocate resources effectively and intervene in a timely manner. The integration of machine learning and AI in these systems has led to significant advancements in predictive healthcare.

\subsection{Machine Learning in Healthcare Prediction}
Gong et al.~\cite{Gong2021} developed a machine learning-based framework for predicting acute kidney injury (AKI), showcasing an end-to-end decision support system that encompasses data pre-processing, risk prediction, and model explanation. This framework utilized logistic regression, random forest, and a voting-based ensemble model, along with gradient boosting algorithms, to address the challenges posed by imbalanced datasets. The model's prediction capability within 48 hours was complemented by SHapley Additive exPlanations (SHAP) values for a dual perspective: a global view highlighting critical factors and a local view detailing individual patient-level feature contributions. In addition, Wu et al.~\cite{Wu2018} compared eight feature selection methods to enhance AKI prediction, underlining the importance of feature selection stability and similarity.

\subsection{Assessment of Interpretability Techniques}
ElShawi et al.~\cite{ElShawi2020} proposed quantitative measures to assess the quality of several model-agnostic interpretability techniques, including LIME, SHAP, Anchors, and others. Their study utilized a random forest model to predict mortality and diabetes risk, evaluating the performance of these interpretability techniques in terms of similarity, bias detection, execution time, and trust. In a separate study, Elshawi et al.~\cite{ElShawi2019} applied global and local explainability techniques to predict the risk of hypertension, enhancing the transparency of machine learning outcomes. Ilic et al.~\cite{ilic2021explainable} introduced an explainable boosted linear regression (EBLR) algorithm for time series forecasting, demonstrating that maintaining interpretability does not necessarily compromise model performance.

\subsection{Attention Mechanism in Deep Learning}
The attention mechanism, initially a breakthrough in machine translation tasks, has been adapted for healthcare applications. Bari et al.~\cite{Bari2021} conducted an empirical evaluation of attention-based deep neural networks, assessing prediction performance, explainability correctness, and sensitivity. Their results indicated that multi-variable LSTM models with explainability features performed well with complex data. Kaji et al.~\cite{Kaji2019} implemented an attention-based LSTM model for predicting medical conditions like sepsis and myocardial infarction, using MIMIC-III dataset patient data. They highlighted the importance of the attention layer in extracting influential input features for better explainability. Chen et al.~\cite{chen2023interpretable} further advanced this field by proposing bilateral asymmetric skewed Gaussian attention (bi-SGA) to improve the performance and interpretability of deep convolutional neural networks.

\subsection{Gap in Literature and Study Contribution}
The literature reveals that while deep learning is capable of predicting vital signs with minimal healthcare domain knowledge, its lack of explainability remains a significant drawback. This underscores the need for explainable AI methods to demystify the results produced by these "black-box" models. Our study addresses this gap by introducing a novel framework that not only estimates feature importance for enhancing explainability but also provides both global and local interpretations of deep learning model predictions. This comprehensive framework aims to balance the trade-off between deep learning model performance and its explainability, thereby contributing significantly to the field of predictive healthcare.

\section{Research Problem}\label{problem}
The central research problem tackled in this study is the elucidation of deep learning model results, particularly the interpretation of predictions based on independent feature inputs in healthcare settings. This task involves comprehending the causal relationships between input factors and their effect on model predictions. It’s crucial for healthcare professionals to grasp the rationale behind AI-driven predictions, understanding how variations in input feature values can influence these predictions. In a scenario where a deep learning model $M$ uses $N$ features, denoted as $x_j$ (where $j = 1, …, N$), to predict an output $y$, the research aims to elucidate how each input feature $x$ contributes to this prediction. This understanding is vital for models where weights $w_j$ are applied to respective features $j$ at different layers of model $M$. This process can be mathematically represented as:

\begin{equation}\label{rp}
y \longleftarrow f_{M}(w_j \cdot x_j)
\end{equation}

To enhance the explainability of predictions from complex, non-linear models such as neural networks and deep learning, it is essential to quantify the contribution of each feature, $\varphi {x{j}}$, in a comprehensible manner. To enhance the explainability of non-linear model predictions, the contribution of each feature $\varphi _{x_{j}}$ can be estimated into two patterns. 

\begin{equation}\label{contribution}
    {\displaystyle \varphi _{x_{j}} = w_j*x_j - E(w_j*X_j)}
\end{equation}

\begin{equation}\label{contribution2}
    {\displaystyle \sum _{j=1}^{N} \varphi _{x_{j}} = \sum _{j=1}^{N} w_j*x_j - E(w_j*x_j)}
\end{equation}

\begin{itemize}
    \item The first pattern estimates the model output with each feature and subtracts the output with the average effect of all the features, $E(w_j*X_j)$ as shown in Equation~\ref{contribution}. The same approach can estimate the contributions of all features. Summing up all the features' contribution in a prediction instance is, where Equation~\ref{contribution2} shows the predicted value $f_{M}(x)$ minus the average predicted value $E(f_{M}(x))$ for the instance $x$.

    \item The second pattern adds an attention layer to the non-linear model and enables the model to focus on certain important features contributing to the output. This pattern creates a representation $hj$ with $j = 1,…, N$ of each input in vector space, and the weighted sum of the representation act as context vectors as shown in Equation~\ref{context_vector_rp}. Extracting the weights for each input feature can influence output feature contribution $\varphi_{x_{j}}$. 
\end{itemize}
\begin{equation}\label{context_vector_rp}
{\displaystyle c=\sum _{j=1}^{N} \alpha_{j}h_{x_{j}}}
\end{equation}

In this current study, the two patterns estimate feature contribution to explain the prediction process of the deep learning model.

\begin{figure*}
    \centering
    \includegraphics[width=\textwidth]{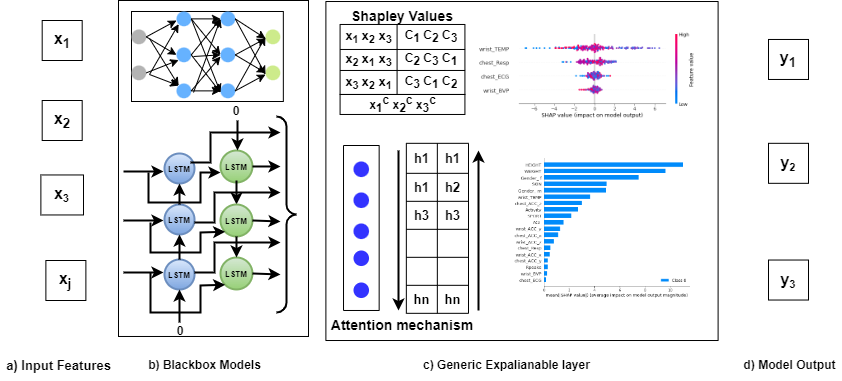}
    \caption{Explainable AI (QXAI) framework}
    \label{fig:qef}
\end{figure*}

\section{Explainable AI for Quantitative data (QXAI)}\label{methods}
In this section, Explainable AI for Quantitative data (QXAI) is proposed to estimate input feature importance in deep learning model results that could be prediction or classification tasks. The proposed framework can provide explainability at two levels, one is post-hoc explainability using Shapley values and the other is intrinsic explainability using attention mechanism as shown in Fig.~\ref{fig:qef}.

\subsection{Shapley Values Calculation}
To explain the contribution of input features, the Shapley value concept based on a coalition game was adopted~\cite{lundberg2017unified}. The coalition game theory can be defined by designating a value for each coalition game with a limited set of players $N$, $S \subseteq N$ to be a subset of $|S|$ players and a characteristic function ${v:2^{N}}\to \mathbb{R}$ from the set of all possible coalitions of players to a set of players that satisfies ${v}(\emptyset)=0$ where $(\emptyset)$ is an empty set. This function determines each player's contribution to the outcome, and the game can be called a profit game or value game. 

The profit game or value game can be adapted to the proposed QXAI framework to determine players (features) contributing to the prediction capacity of a trained deep learning model. To attribute a value to the contribution of each feature, the Shapley value concept can be adapted to explain the contribution in terms of expected marginal contribution. Shapley values assume that all the features contribute to the outcome, and the amount that each feature $x_{j}$ contributes in a coalition game $(v, N)$ is shown in Equation~\ref{shap}. 

\begin{equation}\label{shap}
    {\displaystyle \varphi _{x_{j}}(v)=\sum _{S\subseteq N\setminus \{x_{j}\}}{\frac {|S|!\;(n-|S|-1)!}{n!}}(v(S\cup \{x_{j}\})-v(S))}
\end{equation}

where the sum extends over all subsets S of N not containing feature i and n is the total number of features.

The above Equation~\ref{shap} can further break-down to have individual feature contribution as $v(S\cup{x_{j}})-v(S)$. The characteristic function $v(S)$ can be calculated by using Kernel SHAP.

\begin{equation}\label{indi_shap}
    {\displaystyle \varphi _{x_{j}}(v)={\frac {1}{n!}}\sum _{R}\left[v(P_{x_{j}}^{R}\cup \left\{x_{j}\right\})-v(P_{x_{j}}^{R})\right]}
\end{equation}

where the sum iterates over all ${\displaystyle n!}$ orders ${\displaystyle R}$ of the features and ${\displaystyle P_{x_{j}}^{R}}$ is the set of features in ${\displaystyle N}$ which proceeds the order ${\displaystyle R}$.

In simple terms, Shapley of a feature $x_{j}$ can be defined as below, Equation~\ref{simple}:

\begin{equation}\label{simple}
    {\displaystyle \varphi _{x_{j}}(v)={\frac {1}{n}\sum _{K}{\frac {\varphi(x_{j})}{Z}}}}
\end{equation}

Where $n$ is a number of features, $\varphi(x_{j})$ is marginal contribution of feature $x_{j}$ to coalition, $K$ is coalitions excluding $x_{j}$, $Z$ is a number of coalitions excluding $x_{j}$.

Shapley proposed four conditions (or axioms) below that must be satisfied to have fair contribution of features to a prediction. Equations~\ref{shap},\ref{indi_shap} obey these conditions while estimating the contribution value of each feature.

\begin{itemize}
    \item The summation of Shapley values of all agents equals the value of the total coalition.
    \item All features have a fair chance to participate in a prediction outcome by including in all permutations and combinations of the features.
    \item If a participated feature $x_{j}$ contributes nothing to a prediction outcome, then zero value is attributed to the feature's contribution.
    \item For any pair of predictions $v,w:\varphi(v+w)=\varphi(v)+\varphi(w)$ in which the values are based on additive property $(v+w)=v(S)+w(S)$ for all subsets $S$.
\end{itemize}

\subsection{Kernel SHAP}
Kernel SHAP is a model-agnostic method from the combination of classical Shapley values discussed in Equations~\ref{shap},\ref{indi_shap} and local explainable model-agnostic explanations (LIME) to approximate SHAP values. Instead of retraining models with a subset of features $|S|$, the full model $f$ can be used which is already trained, while replacing missing features with marginalized features. Considering an instance with three features $x_1,x_2,x_3$ and following Equation~\ref{kernel} estimates a partial model with $x_3$ being missed. However, $p(x_3)$ is still required to approximate the missing $x_3$ feature. To address this, a custom proximity function $\pi$ from LIME as shown in Equation~\ref{custom-prox} and SHAP similarity kernel equation~\ref{custom-shap} can be used.
\begin{equation}\label{kernel}
    {\displaystyle f_{x_1,x_2}(x_1,x_2) {\longrightarrow} \int{f(x_1,x_2,x_3)p(x_3)dx_3}}
\end{equation}

\begin{equation}\label{custom-prox}
    {\displaystyle \pi_{x}^{LIME} (z) = \exp(-D(x,z)^2/\sigma^2)}
\end{equation}

\begin{equation}\label{custom-shap}
    {\displaystyle \pi_{x}^{SHAP} (z') = \frac{(p-1)}{\binom{p}{|z'|}|z'|(p-|z'|)}}
\end{equation}

Equation~\ref{custom-prox} penalizes the distance between sample points and the original features' data, for which explainability is being estimated. In Equation~\ref{custom-shap}, coalitions with a number of features that are far from 0 and $p$ will be penalized. The equation adds more weight to coalitions with a small set of features or almost all the features to highlight the independent behavior of the feature or the impact of the features in interaction with others. The choice of this SHAP similarity kernel is based on three properties of additive feature attribution methods local accuracy, missingness, and consistency~\cite{lundberg2017unified}. In this study, Kernel SHAP is used to estimate the contributions of each feature $x_{j}$ value to the prediction. It consists of five steps: 1) Sample coalitions with features and without features. 2) Get prediction for each sample coalition by first converting to the original feature space and applying the machine learning model. 3) Estimate the weight for each coalition with the SHAP kernel. 4) Fit the weighted linear model. 5) Return Shapley value $\varphi _{x_{j}}(v) $ and the coefficients of the model.

\subsection{Attention Mechanism}
The attention mechanism is a widely adopted concept in Natural Language Processing (NLP) tasks like neural machine translations and extracting the cause-effect of input features to model output~\cite{vaswani2017attention,Bahdanau}. The attention mechanism predicts the outcome with better accuracy because its cognitive capability can enhance certain parts of important input data for deep learning model training. The idea of using the attention mechanism to model explainability is to identify the weights beings assigned to each input feature in predicting the outcome. This assists in decoding the importance of each feature and enables human explanation of the cause-effect of the input features. 

An attention layer added to a deep learning model can mimic the cognitive capability of the attention mechanism. Given a set of input features $N$,  $x_j$ is a feature value, with $j = 1,…, N$ to predict an output value $y$. A Bidirectional Long Short-Term Memory (BiLSTM) model can generate vector representations of the input features, such as $h_j$ with, $j = 1,…, N$ based on the forward and backward hidden states in the deep learning model. A generic encoder-decoder model focuses on the last state of the encoder LSTM model and uses it as a context vector. This would cost the information loss of previous states. Attention acts as an interface between the encoder and decoder states of the BiLSTM model and provides a context vector to the decoder with information from every encoder's hidden states. For each prediction value $y$, a context vector $c$ is generated using the weighted sum of the vector representations, as shown in Equation~\ref{context_vector}. The weights $\alpha_{j}$ are computed using a softmax function as shown in Equation~\ref{softmax}. The output score $e_{j}$ is calculated in a feedforward neural network described by a function $f$ to capture alignment between input feature $x_{j}$ and output $y$. The input features are then multiplied (dot product) with $(w_{j}+B)$ where $w_{j}$ is weight and $B$ bias followed by a tan hyperbolic function to estimate the score $e_{j}$ as shown in Equation~\ref{alignment_score}

\begin{equation}\label{context_vector}
{\displaystyle c=\sum _{j=1}^{N} \alpha_{j}h_{x_{j}}}
\end{equation}

\begin{equation}\label{softmax}
{\displaystyle \alpha_{j} = softmax(e_{j})=\frac{exp(e_{j})}{\sum_{j=1	}^{N} exp(e_{j})}}
\end{equation}

\begin{equation}\label{alignment_score}
{\displaystyle e_{j} = f(x_{j},h_{x_{j}}) = tanh(x{j} \cdot (w_{j}+B)})
\end{equation}

For input features $x_{1},x_{2},x_{3},x_{4}$ , let the weights $\alpha_{j}$ be, $[0.2,0.4,0.6,0.1]$ then the context vector would be as shown in equation~\ref{context_vector_example}. This can assist in estimating the importance of each input feature in the context vector, which will be fed to the decoder network for model predictions.
\begin{equation}\label{context_vector_example}
{\displaystyle c= 0.2 \times x_{1} + 0.4 \times x_{2} + 0.6 \times x_{3}+ 0.1 \times x_{4}}
\end{equation}

\subsection{Global and Local explanation}
Two different forms of explanation perspectives such as global explanation and local explanation are proposed in this study. The global explanation can provide the contribution of each feature in the prediction of vital sign. This is designed to assist clinicians by providing holistic information about the prediction and to identify which clinical factors or features need special attention. To estimate the global importance of the features in the prediction, the absolute Shapley values calculated from Equation~\ref{shap} are averaged for each feature across the data, as shown in Equation~\ref{GI}. Based on this calculation, the features can have their importance sorted in descending order. 

\begin{equation}\label{GI}
    {\displaystyle I_i=\frac{1}{n} \sum_{i=1}^{n}|\varphi_{i}|}
\end{equation}

Although feature importance can provide an overview of all selected features' importance towards a prediction, it cannot uncover the correlation of the features with a target variable and estimate contributing and non-contributing data points of a feature. This, however, can be achieved by using Shapley values of each feature on a summary plot showing the level of positive and negative contribution to a target variable.  

In the case of local explanation, vital signs prediction at each time step can be decrypted. This can summarize features that are aiding the patient's health in terms of vital signs and can enable personalized monitoring, which is critical in healthcare applications. The Shapley values of each feature can be positive or negative, and each value is considered a force that either increases or decreases the prediction value. This helps to explain individual features that are forcing the prediction value to either increase or decrease. The local explanation concept can be applied to an individual record in a prediction or a group of records related to a specific subject or activity. 
\begin{algorithm} 
\caption{Feature contribution estimation}
\begin{algorithmic}[1]\label{algorithm1}
\REQUIRE{a set of features $\mathcal{F}=\{1,2,\dots,N\}$};{a set of deep learning models $\mathcal{M}=\{m_{1},m_{2}\}$ where $m_{1}$ is without attention and $m_{2}$ is with attention};{a input dataset D}\vfill
\ENSURE{Contributions of the features $\mathcal{F}=\{1,2,\dots,N\}$} in the form of Shapley values and attention weights;\vfill

\STATE Split dataset: $D=D^{train} \vee D^{test}$

{\textbf{Global explanation}}
\STATE ${m_{1}^{train}\longleftarrow D^{train} }$
\STATE ${m_{1}^{test}\longleftarrow D^{test}}$
\STATE $Shapley\_values\leftarrow kernelshap({m_{1}^{train}}, D^{test})$

\STATE ${m_{2}^{train}\longleftarrow D^{train} }$
\STATE ${m_{2}^{test}\longleftarrow D^{test}}$
\STATE $attention\_weights\leftarrow model.attention\_weights()$

{\textbf{Local explanation}}

\FOR {d in D}
    \STATE $Shapley\_values\leftarrow kernelshap({m_{1}^{train}}, d)$
    \STATE $attention\_weights\leftarrow m_{2}.attention\_weights()$
\ENDFOR
\end{algorithmic}
\end{algorithm}

\subsection{QXAI Algorithm}
The proposed QXAI framework comprises two deep learning model approaches, one with model attention and the second without. The framework can be implemented with the Algorithm~\ref{algorithm1} and can be adapted to execute global and local explanations. In Algorithm~\ref{algorithm1}, line 1 splits the input data into test and train sets to train and evaluate the deep learning models. Lines 2-7 present the global explanation using kernel SHAP and attention layer weights. Lines 2-4 train a deep learning model without an attention layer and pass it to the kernel SHAP explainer to extract Shapley values of the input features. Lines 5-7 present the attention-based deep learning model and extracts the attention layer weights, thus defining input feature importance. Lines 8-11 present the local explanation for each input record d from data D. 

\begin{figure*}[ht]
    \centering
    \includegraphics[width=0.7\textwidth]{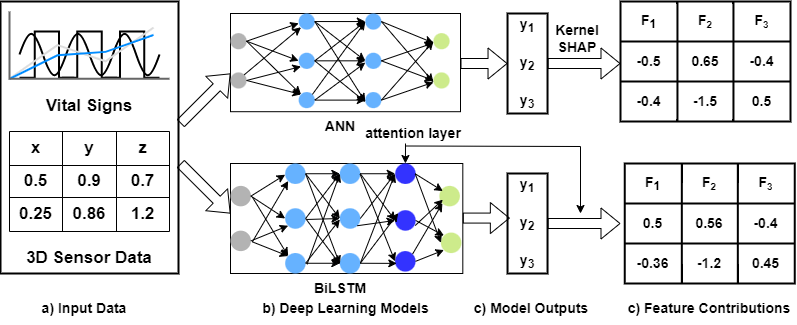}
    \caption{Experimental Design}
    \label{fig:design}
\end{figure*}

\section{Experiment}\label{exp-design}
The two key aspects of an explainable AI framework are the understanding phase and the explaining phase~\cite{dwivedi2023explainable}. The former is concerned with improving models during training by interpreting critical features and building robust models, while the latter involves deploying and providing human-readable explanations to end users. Striking a balance between model performance and explainability is always a challenge in AI applications. In AI applications, there is always a trade-off between model performance and explainability~\cite{gunning2019darpa}. According to Zacharias et al.~\cite{zacharias2022designing}, the preprocessing stage, specifically feature selection, has been overlooked in explainable AI applications and requires attention. The importance of each feature to the outcome can be used for semantic labeling and to improve cognitive understanding, as it provides positive framing and direction (positive or negative contribution).

To address the limitations of explainable AI, the proposed QXAI framework in this study focuses on feature selection and provides local and global explanations through post-hoc models and intrinsic weights. The study evaluates feature importance in the QXAI framework, which can reduce dimensionality, improve cognitive understanding, and help with decision-making. Good explanations are crucial for making informed decisions, especially in dynamic domains like healthcare. In addition to the feature importance step in explainability, this study further breaks down the explainability into local and global explanations for each supervised learning task in classification and regression. Local and global explanations help in understanding the positive or negative contribution of features at the model and individual levels. The study used publicly available benchmark datasets for evaluation. Figure~\ref{fig:design} illustrates the experimental design of the proposed framework. 


\subsection{Datasets}

\begin{itemize}
    \item \textbf{PPG-DaLiA}~\cite{reiss2019deep}: This dataset from 15 subjects comprised physiological and motion data while performing a wide range of activities under close to real-life conditions. The collected data were from both a wrist-worn (Empatica E4) and a chest-worn (RespiBAN) device. The dataset consists of 11 attributes, including 3-dimensional (3D) acceleration data, electrocardiogram (ECG), respiration, blood volume pulse (BVP), electrothermal activity (EDA), and body temperature. 
    
    
    \item \textbf{MHEALTH}~\cite{Banos2014}: This dataset comprises the body motion of ten volunteers while performing 12 physical activities recorded from three sensors at the chest, left ankle, and right lower arm. There were 21 independent attributes including acceleration, gyroscope, and magnetometer of the three sensors. A dependent variable classifying the 12 activities was based on the sensor data.
    
    
    \end{itemize}

\begin{table*}[ht]
\centering
\caption{Implementation details}
\label{tab:impledet}
\resizebox{0.7\textwidth}{!}{%
\begin{tabular}{@{}ccc|cc@{}}
\toprule
\multicolumn{1}{l}{} &
  \multicolumn{2}{c}{\textbf{Regression}} &
  \multicolumn{2}{c}{\textbf{Classification}} \\ \midrule
\multicolumn{1}{l}{} &
  \multicolumn{1}{c}{\textbf{\begin{tabular}[c]{@{}c@{}}Shapley \\ Values\end{tabular}}} &
  \multicolumn{1}{c}{\textbf{\begin{tabular}[c]{@{}c@{}}Attention \\ Mechanism\end{tabular}}} &
  \multicolumn{1}{c}{\textbf{\begin{tabular}[c]{@{}c@{}}Shapley \\ Values\end{tabular}}} &
  \multicolumn{1}{c}{\textbf{\begin{tabular}[c]{@{}c@{}}Attention \\ Mechanism\end{tabular}}} \\ \midrule
\multicolumn{1}{c}{\textbf{Models}} &
  \multicolumn{1}{c}{ANN} &
  \multicolumn{1}{c}{BiLSTM} &
  \multicolumn{1}{c}{ANN} &
  \multicolumn{1}{c}{BiLSTM} \\ \midrule
\multicolumn{1}{c}{\textbf{No of Layers}} &
  \multicolumn{1}{c}{5} &
  \multicolumn{1}{c}{4+attention layer} &
  \multicolumn{1}{c}{5} &
  \multicolumn{1}{c}{4+attention layer} \\ \midrule
\multicolumn{1}{c}{\textbf{\begin{tabular}[c]{@{}c@{}}Activation\\ Functions\end{tabular}}} &
  \multicolumn{1}{c}{\begin{tabular}[c]{@{}c@{}}relu, \\ sigmoid\end{tabular}} &
  \multicolumn{1}{c}{\begin{tabular}[c]{@{}c@{}}relu, \\ Softmax\end{tabular}} &
  \multicolumn{1}{c}{\begin{tabular}[c]{@{}c@{}}LeakyReLU, \\ Sigmoid\end{tabular}} &
  \multicolumn{1}{c}{\begin{tabular}[c]{@{}c@{}}relu, \\ Softmax\end{tabular}} \\ \midrule
\multicolumn{1}{c}{\textbf{Optimizers}} &
  \multicolumn{2}{c}{Adam} &
  \multicolumn{2}{c}{Adam} \\ \midrule
\multicolumn{1}{c}{\textbf{loss Functions}} &
  \multicolumn{2}{c}{mean\_absolute\_error} &
  \multicolumn{2}{c}{binary\_crossentropy} \\ \midrule
\multicolumn{1}{c}{\textbf{Epochs}} &
  \multicolumn{2}{c}{100} &
  \multicolumn{2}{c}{100} \\ \midrule
\textbf{Batch Size} &
  \multicolumn{2}{c}{64} &
  \multicolumn{2}{c}{64} \\ \bottomrule
\end{tabular}}
\end{table*}

\subsection{Data Modelling}
Datasets consisted of preprocessed raw data from the sensor's signal and features were stored in different CSV files. In this step of data preparation, the dataset was further preprocessed to have a single structured file with a set of features for each subject. The datasets were prepared for two different tasks: regression and classification. The regression task was to predict the heart rate of the subjects based on their sensor readings. The classification task was to classify the physical activities of the subjects based on their motion data recorded from three axes of sensors. The physical activities label was preprocessed to have a multi-label classification. Each of these datasets was split into an 80:20 ratio for 80\% of data for training and 20\% of data for testing. 

In this study, two deep learning models artificial neural networks (ANN), and Bidirectional LSTM (BiLSTM) models were adopted. The ANN model was configured with an input layer, hidden layers, and an output layer. The traditional activation function rectified linear unit (ReLU) has a limitation of defining negative inputs to zero which deactivates the nodes or neurons. Considering the negative values in 3D sensor data, the ANN model used the activation function LeakyReLU in input and hidden layers to avoid the zero input values of the negative attributes. The output layer was configured with the traditional activation function ReLU to predict the target variable heart rate greater than zero based on the activation function property. The loss function used for the regression study was mean absolute error, which also acted as a performance metric for the model. For the classification task, binary cross entropy acted as a loss function along with metrics like accuracy.  The Adam adaptive optimizer~\cite{kingma2014adam} was chosen for the model for its quick computational time, it requires fewer parameters for tuning compared to other optimizers. The attention mechanism discussed in the proposed framework was added to the BiLSTM model, which has encoder and decoder states to generate vector representations. The preprocessed data was fed to the attention-based BiLSTM model and extracted the attention layer weights. This determined the input feature importance in the deep learning model prediction. 

The datasets in this study were created by preprocessing raw data from sensor signals and storing the features in separate CSV files. These datasets were then combined into a single structured file for each subject, with separate datasets prepared for regression and classification tasks. The regression task involved predicting the subject's heart rate based on sensor readings, while the classification task involved categorizing the subject's physical activities using motion data from three axes of sensors. The datasets were split into 80\% for training and 20\% for testing. Two deep learning models, ANN and BiLSTM, were used in this study as shown in Tab.~\ref{tab:impledet}. The table presents implementation details of ANN and BiLSTM models in regression and classification tasks. 

For regression tasks, the models use the Shapley Values and attention mechanism. The ANN model has 5 layers with the activation functions of relu and sigmoid. The BiLSTM model has 4 layers with an additional attention layer with the activation functions of ReLU and softmax. The optimizer used is Adam, and the loss function is mean\_absolute\_error. For classification tasks, the models also use ANN and BiLSTM architectures, Shapley Values, and attention mechanisms. The ANN model has 5 layers with the activation functions of LeakyReLU and sigmoid. The optimizer used is Adam, and the loss function is binary\_crossentropy. The BiLSTM model has 4 layers with an additional attention layer. The activation functions used are ReLU and softmax. For both prediction and classification tasks, the models are trained for 100 epochs with a batch size of 64.



\subsection{Traditional Models}
By comparing the feature importance estimated using Shapley values and intrinsic weights of the attention mechanism with the traditional machine learning models, the explainability of the proposed framework was evaluated. The two deep learning models in the framework, ANN and BiLSTM, were also evaluated to ensure high performance and robustness with explainability. This allowed the study to evaluate the effectiveness of the framework in explainability without compromising model performance.

The proposed approach was evaluated with models with state-of-art performances. The deep learning models adopted in the proposed approach were compared with heart rate prediction and human activity recognition performances. The feature importance was compared with traditional machine learning models, which had the capability to produce feature importance for prediction and classification results. \\
\textbf{Prediction}
\begin{itemize}
    \item Ni et al.~\cite{ni2019modeling} proposed context-aware sequential models to capture personalized fitness data and forecast heart rate to recommend suitable activities. The authors used a multi-layer perceptron model to forecast heart rate. 
    \item Zhu et al.~\cite{zhu2022fitness} proposed four LSTM models for an optimization training system to predict heart rate under three different types of exercises walking, rope jumping, and running. Three of the four LSTM models were used for heart rate prediction and one for human activity recognition.
\end{itemize}
\textbf{Classification}
\begin{itemize}
    \item In a previous study, we proposed FedStack~\cite{Shaik2022}, a novel federated framework to classify patients' physical activities. We adopted deep learning models such as CNN, ANN, and BiLSTM for the classification.
    \item Bozkurt et al.~\cite{Bozkurt2021} compared deep learning model performance with traditional machine learning models for human activity recognition. Deep Neural Network (DNN) model achieved an accuracy of 96.81\% and outperformed other models.
\end{itemize}
\textbf{Feature Importance}
\begin{itemize}
    \item Li et al.~\cite{yijing2022prediction} proposed an explainable machine learning model named cardiac arrest prediction index for early detection of cardiac arrest. The authors used the XGBoost model for the prediction and achieved an area under the receiver operating characteristic curve (AUROC) of 0.94. 
    \item Gong et al.~\cite{Gong2021} used XGBoost and voting ensemble method combining random forest and logistic regression to predict acute kidney injury. For explanation, the SHAP technique was used to understand important predictors and relationships among the predictors.
    \item Ali et al.~\cite{Ali2021} proposed supervised machine learning algorithms such as Random Forest, Decision Tree, and KNN for heart disease prediction. Feature importance scores for each feature were computed with Decision Tree and Random Forest \cite{malakar2022computer}. 
\end{itemize}

\begin{figure*}[!ht]
    \centering
    \begin{subfigure}[b]{0.45\textwidth}
        \centering
        \includegraphics[width=\textwidth]{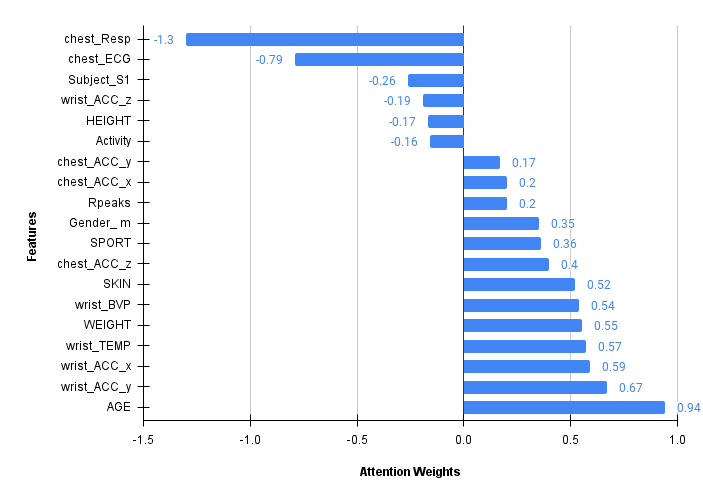}
        \caption{Attention Weights}
        \label{fig:atten_weights_pred}
    \end{subfigure}%
    \hfill
    \begin{subfigure}[b]{0.45\textwidth}
        \centering
        \includegraphics[width=\textwidth]{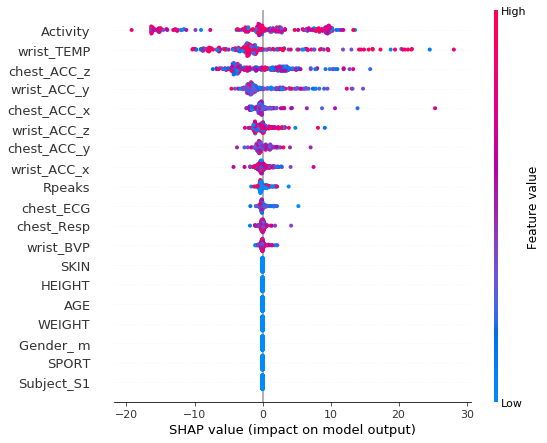}
        \caption{Shapley Values}
        \label{fig:shap_pred}
    \end{subfigure}
    \newline
    \begin{subfigure}[b]{0.45\textwidth}
        \centering
        \includegraphics[width=\textwidth]{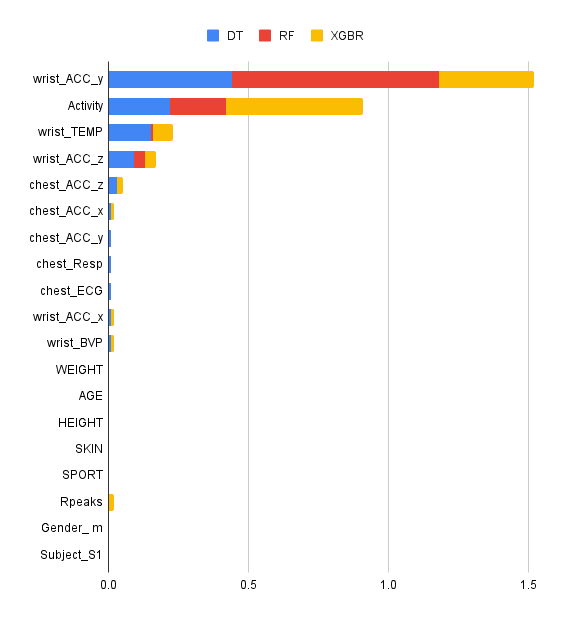}
        \caption{Traditional Models}
        \label{fig:baseline_pred}
    \end{subfigure}
    \caption{Regression Model—Feature Importance Plots}
    \label{fig:combined_figures}
\end{figure*}

\subsection{Performance Metrics}
Explainability is a multifaceted concept, and there is no single metric to measure it. The evaluation of explainability involves comparing the feature importance provided by different models, such as comparing the explanations of ANN and BiLSTM with those of traditional models. In this study, another two sets of performance metrics were adapted to evaluate deep learning models' prediction and classification results. For prediction, mean absolute error (MAE) and mean squared error (MSE) was used to evaluate the performance of the prediction model. Both metrics measure the deviation or difference of a predicted value from its actual value. For classification, a traditional confusion matrix was used to calculate precision, F1-Score, recall, and balanced accuracy metrics of deep learning results on multi-label classification.

\section{Experimental Results and Analysis}\label{results}
In this section, we analyze the evaluation results of the proposed QXAI framework. The results are focused on explainability in terms of feature importance for positive framing, local explanations for semantic labelling to explain the positive or negative contributions of each input feature to the deep learning model's prediction, and global explanations that can explain a model's overall predictions with interactive plots. To address the trade-off between explainability and model performance in AI applications~\cite{herm2023stop}, the performance of the deep learning models, ANN and BiLSTM-attn, in the framework for both regression and classification tasks was evaluated and compared with those of traditional machine learning models.


\begin{table}[!ht]
\scriptsize
\centering
\caption{QXAI Prediction Performance}
\label{tab:global_perf}
\begin{tabular}{@{}ccc@{}}
\toprule
\textbf{Model}                                     & \textbf{MAE}                       & \textbf{MSE}                        \\ \midrule
\multicolumn{1}{c}{ANN}            & \multicolumn{1}{c}{\textbf{3.33}} & \multicolumn{1}{c}{\textbf{24.51}} \\ \midrule
\multicolumn{1}{c}{BiLSTM-attn}            & \multicolumn{1}{c}{\textbf{4.40}} & \multicolumn{1}{c}{\textbf{43.72}} \\ \midrule
\multicolumn{1}{c}{MLP~\cite{ni2019modeling}}  & \multicolumn{1}{c}{4.71} & \multicolumn{1}{c}{47.95} \\ \midrule
\multicolumn{1}{c}{LSTM~\cite{zhu2022fitness}} & \multicolumn{1}{c}{5.54} & \multicolumn{1}{c}{69.03}  \\ \bottomrule
\end{tabular}
\end{table}
\subsection{QXAI in Regression Problem}
The proposed QXAI approach was evaluated on its ability to predict heart rate based on sensor data and clinical indicators. Other vital signs retrieved from human subjects were in the PPG-DaLiA dataset. The two deep learning models ANN and attention-based BiLSTM proposed in the framework were trained on the data to predict the vital signs. The models' performance was compared with other traditional models shown in Tab.~\ref{tab:global_perf}. The ANN model performed better than the attention-based BiLSTM, MLP, and LSTM models with MAE and MSE of 3.33 and 24.51 respectively. 


\begin{figure*}[!ht]
    \centering
    \begin{subfigure}[b]{0.9\textwidth}
        \centering
        \includegraphics[width=\textwidth]{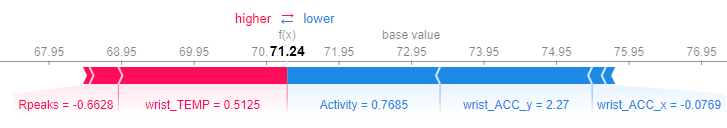}
        \caption{Prediction—Local explanation}
        \label{fig:local_interpret}
    \end{subfigure}
    \newline
    \begin{subfigure}[b]{0.9\textwidth}
        \centering
        \includegraphics[width=\textwidth]{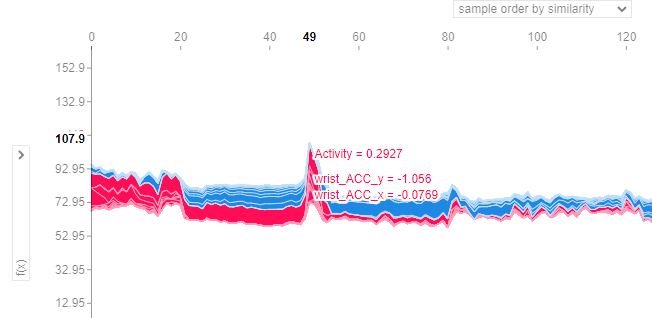}
        \caption{Prediction—Global explanation}
        \label{fig:Group_interpret}
    \end{subfigure}
    \caption{Explanations for prediction: (a) Local explanation illustrating individual feature contributions. (b) Global explanation showing overall feature contributions.}
    \label{fig:explanations_combined}
\end{figure*}

\subsubsection{Feature Importance}
Feature importance of input features was estimated using the proposed QXAI approach and compared with traditional machine learning model feature importance. The three feature importance plots shown in Fig.~\ref{fig:atten_weights_pred}, \ref{fig:shap_pred},and~\ref{fig:baseline_pred} present attention weights retrieved from the BiLSTM model, Shapley values estimated from Kernel SHAP, and traditional model feature importance respectively. The y-axes in each subplot hold the input features, with x-axes showing the importance of each feature to the respective model's prediction. The large value of the x-axis determines the importance or contribution of a feature to model performance in predicting heart rate. Activity, chest, and wrist sensors data had high feature importance for heart rate prediction compared to other input vital signs like wrist\_BVP, chest\_Resp, and chest\_ECG. The Shapley values plot~\ref{fig:shap_pred} and attention weights plot~\ref{fig:atten_weights_pred} presented the negative dimensions of each feature's contribution.

\subsubsection{Explainability}
As discussed in Section~\ref{methods}, global and local explanations both contribute to presenting a patient's health status at different levels. The local explanation assists the clinician to explain the health status at a particular time step of patient monitoring.  Fig.~\ref{fig:local_interpret} presents feature contribution to the ANN model label prediction for a selected random record. The randomly selected record is of a male subject aged 25 years, height 168 centimeters, weight 57 kilograms with fitness level 5 on a scale 1-6 where 1 refers to them exercising less than once a month and 6 refers to 5-7 times a week. The subject's activity was measured during his lunch break, and his heart rate prediction was 71.24. The red highlighted features in Fig.~\ref{fig:local_interpret} indicated a negative contribution and pushed the prediction value to the right (higher) side of the scale, whereas the blue features positively contributed and pushed the prediction value to the left (lower) side of the scale. This infers activity, wrist\_ACC\_y, and wrist\_ACC\_x features are negatively contributing and trying to decrease the heart rate value. The Rpeaks and wrist\_TEMP features are balanced by increasing the heart rate to the expected value of 72.95. The SHAP values of each feature can be positive or negative. Similarly, Fig.~\ref{fig:Group_interpret} presents a subject-level explanation of features' contribution to their heart rate prediction based on 200 records. The chart is related to a subject and presents each predicted value on the y-axis with its feature contribution spread on the x-axis in blue and red highlight. This is an interactive plot with dropdowns on the x-axis and y-axis changing and shows the impact of individual features on all 200 predictions. The plot is a screenshot of a prediction value of 107.9 in which the feature activity from wrist\_ACC\_x and wrist\_TEMP are negatively contributing to the heart rate prediction.


\begin{figure}[h]
     \centering
     \begin{subfigure}[b]{\columnwidth}
         \centering
         \includegraphics[width=\columnwidth]{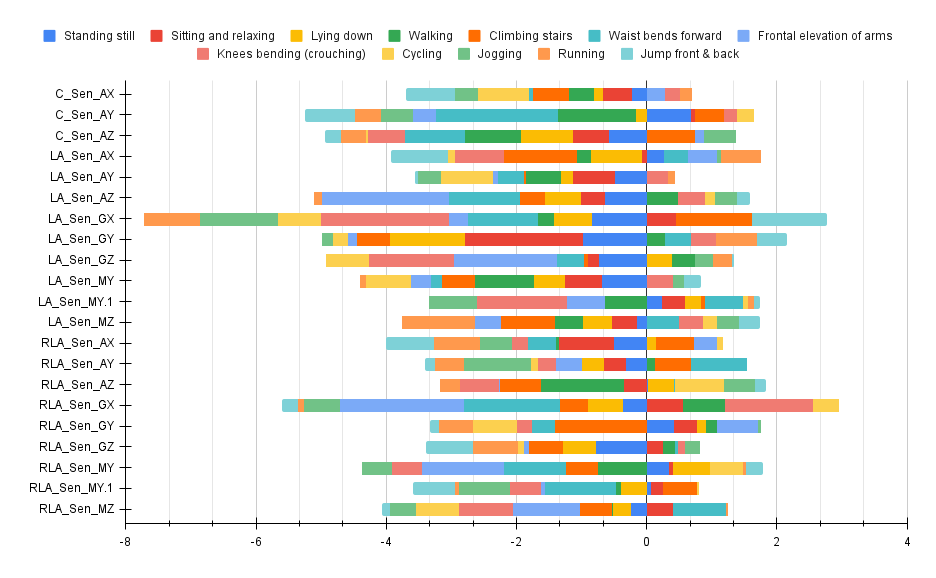}
         \caption{Attention Weights}
         \label{fig:atten_weights_classi}
     \end{subfigure}
     \vfill
     \begin{subfigure}[b]{\columnwidth}
            \centering
            \includegraphics[width=\columnwidth]{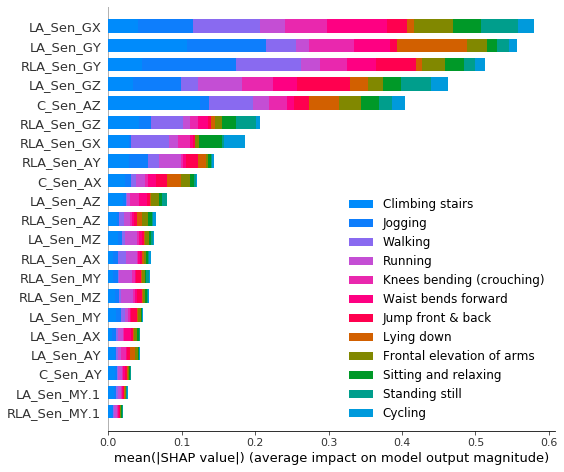}
            \caption{Shapley Values}
            \label{fig:shap_classi}
     \end{subfigure}
     \vfill
     \begin{subfigure}[b]{\columnwidth}
         \centering
         \includegraphics[width=\columnwidth]{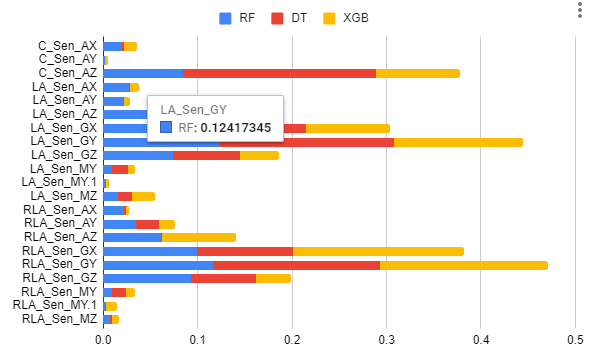}
         \caption{Traditional Models}
         \label{fig:baseline_classi}
     \end{subfigure}
    \caption{Classification Model—Feature Importance Plots}
    \label{fig:FI_classi}
\end{figure}

\begin{table}[!ht]
\centering
\scriptsize
\caption{QXAI Classification Performance}
\label{tab:classifi_results}
\begin{tabular}{@{}lrrrr@{}}
\toprule
\textbf{} &
  \multicolumn{1}{l}{\textbf{Precision}} &
  \multicolumn{1}{l}{\textbf{Recall}} &
  \multicolumn{1}{l}{\textbf{\begin{tabular}[c]{@{}l@{}}F1-\\ score\end{tabular}}} &
  \multicolumn{1}{l}{\textbf{\begin{tabular}[c]{@{}l@{}}Balanced \\ Accuracy\end{tabular}}} \\ \midrule
\textbf{ANN}                                                      & 1    & 1    & 1    & 1    \\ \midrule
\textbf{BiLSTM-Atten}                                             & 0.92 & 0.78 & 0.77 & 0.88 \\ \midrule
\textbf{CNN~\cite{Shaik2022}}                                            & 0.99 & 0.98 & 0.98 & 0.98 \\ \midrule
\textbf{DNN~\cite{Bozkurt2021}}                                            & 0.97 & 0.97 & 0.97 & 0.97 \\\bottomrule
\end{tabular}
\end{table}

\subsection{QXAI in Classification Problem}
The proposed QXAI approach was also used to explain the classification of human physical activities. Both the deep learning models ANN and attention-based BiLSTM were trained on the MHEALTH dataset. Model classification performance was compared to DNN and CNN, as shown in Tab.~\ref{tab:classifi_results}. The ANN model had the best performance, with all evaluation metric values equalling 100\%. CNN and DNN models also performed better than the attention-based BiLSTM model. The proposed framework disclosed the intrinsic weights of each feature in classification and post-hoc model explanations with Shapley values.

\subsubsection{Feature Importance}
The Shapley values and attention weights computed from the deep learning models determined the input feature importance in classifying human physical activities. Feature importance from the deep learning model was compared with feature importance in traditional machine learning models as shown in Fig.~\ref{fig:FI_classi}. The y-axes in all three subplots,~\ref{fig:atten_weights_classi},~\ref{fig:shap_classi},and~\ref{fig:baseline_classi} refer to the 21 input features passed to the deep learning and the x-axes present the importance of a feature to model classification results. The attention-based BiLSTM model assigned more negative weights to all the input features. The sensor attributes at the wrist and ankle area were assigned with more weights in terms of magnitude to classify human physical activities as shown in Fig.~\ref{fig:atten_weights_classi}. The Shapley values plot~\ref{fig:shap_classi} shows full body motion activities such as climbing stairs, jogging, walking, running, and jump front \& back rely on left ankle sensor gyroscope data. The feature importance metrics from traditional machine learning models could not differentiate labels in their plot, as shown in Fig.~\ref{fig:baseline_classi}, but the results show that gyroscope data features contribute more to physical activity classification.


\begin{figure*}[!ht]
    \centering
    \begin{subfigure}[b]{\textwidth}
        \centering
        \includegraphics[width=0.8\textwidth]{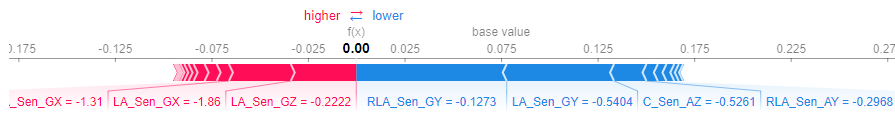}
        \caption{Classification—Local explanation}
        \label{fig:local_interpret_classi}
    \end{subfigure}
    \\
    \begin{subfigure}[b]{\textwidth}
        \centering
        \includegraphics[width=0.8\textwidth]{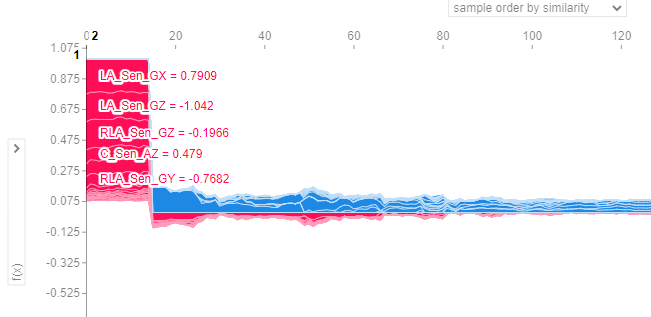}
        \caption{Classification—Global explanation}
        \label{fig:Group_interpret_classi}
    \end{subfigure}
    \caption{Explanations for classification: (a) Local explanation illustrating individual feature contributions. (b) Global explanation showing overall feature contributions.}
    \label{fig:classification_explanations}
\end{figure*}

\begin{figure}[ht]
    \centering
    \begin{subfigure}[b]{\columnwidth}
        \centering
        \includegraphics[width=0.8\textwidth]{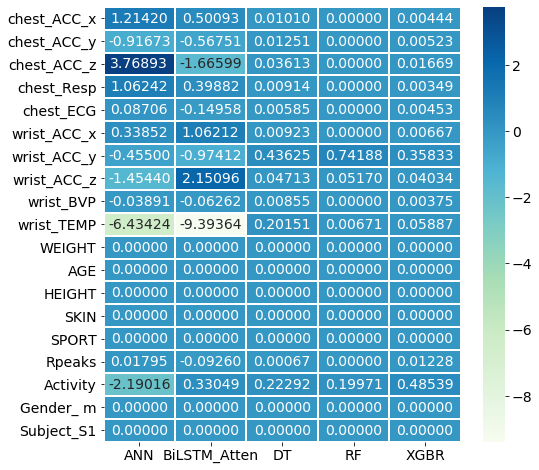}
        \caption{MC Approximation—Prediction}
        \label{fig:mc_pred}
    \end{subfigure}
    \\
    \begin{subfigure}[b]{\columnwidth}
        \centering
        \includegraphics[width=0.8\textwidth]{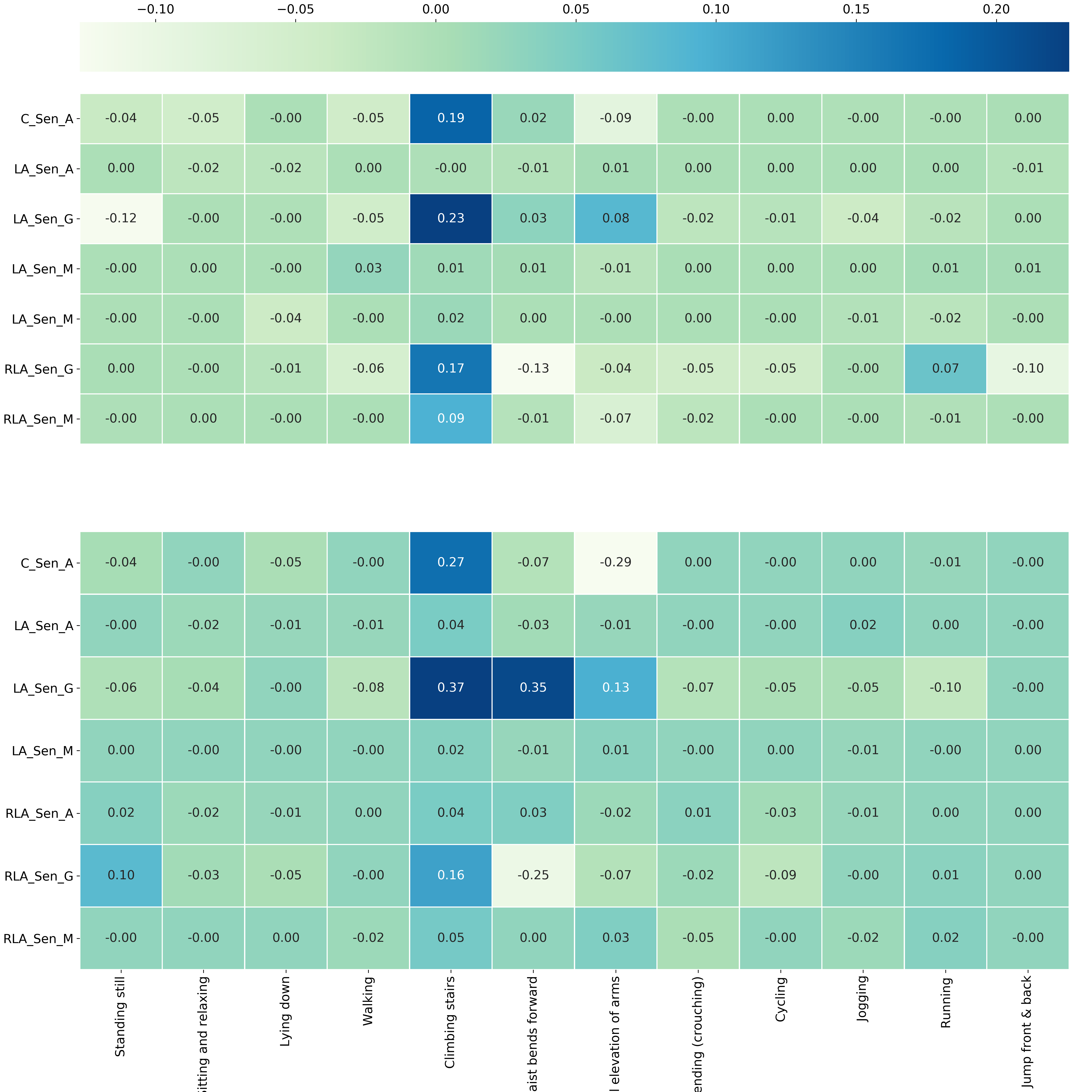}
        \caption{MC Approximation—Classification}
        \label{fig:mc_classi}
    \end{subfigure}
    \caption{Monte Carlo Approximation for Feature Importance Analysis in Prediction and Classification.}
    \label{fig:mc_approximations}
\end{figure}

\subsubsection{Explainability}
The patients' physical activity classification can be explained in detail by breaking down the Shapley values with force plots as shown in Fig.~\ref{fig:local_interpret_classi},~\ref{fig:Group_interpret_classi}. The local explanation at each input record level can assist clinicians to explain physical activity classification and can explain which sensor features are actually contributing to the classification. In plot~\ref{fig:local_interpret_classi}, the ANN model prediction probability of an arbitrarily selected record shows that the x, and z dimensions of the left ankle gyroscope try to push the model probability higher but the y-axis of the right lower arm and left ankle sensors are pushing the probability negatively according to Shapley values' feature importance. Similarly, Fig.~\ref{fig:Group_interpret_classi} presents a subject-level interpretation of features that contribute to their physical activity classification based on 200 records. The chart is related to a subject and presents each predicted value on the y-axis with its feature contribution spread on the x-axis in blue and red highlights. This is an interactive plot with dropdowns on the x-axis and y-axis to change to see the impact of the individual feature on all 200 predictions. The plot is a screenshot of a predicted value 1 in which chest sensor acceleration positively contributes and left ankle and right lower arm sensor features negatively contribute to the heart rate prediction.



\begin{algorithm}
\caption{Monte Carlo Approximation on Feature contribution estimation}
\begin{algorithmic}[1]\label{algorithm2}
\REQUIRE{a set of features $x_{j}=\{1,2,\dots,N\}$};{a set of deep learning models $\mathcal{M}=\{m_{1},m_{2}\}$ where $m_{1}$ is without attention and $m_{2}$ is with attention};{input data $\mathcal{D}$}\vfill
\ENSURE{Contribution of the features $x_{j}=\{1,2,\dots,N\}$}\vfill
 \STATE marginal contribution $\phi_{x_{j}} \leftarrow \emptyset$
\FORALL{$x_{j}=\{1,2,\dots,N\}$}
    \STATE z$\leftarrow$ random sample from $\mathcal{D}$\\
    \STATE x$\leftarrow$ random sample from ${N}$\\
    \STATE choose random permutation o of the feature ${x_{j}}$
    \STATE $x: x_{o}=x_{1},\dots,x_{j}$
    \STATE $z: z_{o}=z_{1},\dots,z_{j}$ \\
    Build two new samples
        \STATE with factor ${F}$:
            \STATE ${x_{+j}= (x_{1},\dots,x_{j-1},z_{o}=z_{1},\dots,z_{j-1})}$
        \STATE without factor ${F}$:
            \STATE ${x_{-j}= (x_{1},\dots,x_{j+1},z_{o}=z_{1},\dots,z_{j+1})}$ \\
    Compute marginal contribution of feature $F$:
        \STATE $\phi_{x_{j}} \leftarrow m_{1}(x_{+j}) - m_{1}(x_{-j})$
\ENDFOR
\STATE $\hat{\phi_{x_{j}}} \leftarrow \frac{1}{x_{j}}\sum_{m=1}^{x_{j}}\phi_{x_{j}}$
\end{algorithmic}
\end{algorithm}

\section{Monte Carlo Approximation}\label{mcApprox}
Feature contributions in model prediction can be estimated based on Shapley value computed using Equation~\ref{shap} proposed in Section~\ref{methods}. These computations have an exponential time complexity and increase in number of features makes the Shapley value calculation unfeasible. In this study, Monte Carlo approximation was adopted to calculate each feature contribution as shown in Equation~\ref{mc-equa}. This approximation technique can extract Shapley values for each feature for both deep learning models. The results have been discussed in this section.

\begin{equation}\label{mc-equa}
    {\displaystyle \varphi _{i} = \frac{1}{N} \sum_{n=1}^{N} (f(d_{+i}^{m}) - f(d_{-i}^{m}) )}
\end{equation}
where $f(\cdot)$ is the contribution of subset features. The $d_{+i}^{m}$ and $d_{-i}^{m}$ is the subset of with and without factor $i$ in subset $n$ features, respectively. 

The implementation of the Monte Carlo approximation is presented in algorithm~\ref{algorithm2}. Lines 3-7 obtain sampled data from the input data D. Lines 8-11 build new samples with or without consideration of a feature $x_{j}$. Line 12 calculates the marginal contribution $\phi_{x_{j}}$ of feature $x_{j}$.  Lines 2-13 are a loop iterating to calculate the contribution of each feature one by one. Finally, line 14 calculates the Shapley value by averaging the outputs of multiple runs.

Monte Carlo approximation was applied to both deep learning models in the proposed QXAI framework. In ANN model prediction results, the approximation technique estimate wrist\_TEMP is the most contributing feature in terms of magnitude, but the negative value shows that the feature is inversely proportional to the heart rate prediction. The other chest\_ACC\_z, chest\_ACC\_x, and chest\_Resp features contributed positively towards the heart rate prediction. The attention weights from the BiLSTM model show that most of the input features are inversely proportional to the model output with negative values. The heat map shows that wrist\_TEMP, wrist\_ACC\_z, and chest\_ACC\_z are the most contributing features to the heart rate prediction as shown in Fig.~\ref{fig:mc_pred}. Similarly, the Monte Carlo approximation was applied to the deep learning classification models. The 3D axes of the sensor inputs were merged to have chest sensor acceleration, left ankle sensors' acceleration, gyroscope, and magnetometer, and right lower arm sensors' acceleration, gyroscope, and magnetometer as shown in Fig.~\ref{fig:mc_classi}. The figure shows ANN model classification Shapley values for the consolidated input feature in the top heat map. The bottom heat map shows the attention-based BiLSTM model Shapley values. The full body activity like climbing stairs classification was more contributed by gyroscope data of the left ankle and right lower arm sensors and acceleration data of the chest sensor.  

\section{Discussion}\label{discussion}

The research presented in this paper makes a significant contribution to the emerging field of explainable AI (XAI) in healthcare, particularly by addressing the challenge of interpretability in deep learning models for vital sign prediction and physical activity classification. The proposed Explainable AI for Quantitative data (QXAI) framework is noteworthy for its innovative approach that combines Shapley values and attention mechanisms, offering a comprehensive dual perspective on both post-hoc and intrinsic explainability. This discussion delves into the implications, strengths, limitations, and future directions of this study.

\textbf{Implications and Contributions:}
The QXAI framework addresses a critical gap in healthcare AI by providing a solution to the 'black-box' nature of deep learning models. This is crucial as the explainability of AI models is increasingly becoming a requirement, especially in high-stakes fields like healthcare. The integration of Shapley values for post-hoc explainability and attention mechanisms for intrinsic understanding allows for a nuanced interpretation of AI decisions. This dual perspective of explainability not only enhances the trustworthiness of AI models but also makes them more practical and useful for clinicians. By enabling healthcare professionals to understand the reasoning behind AI-driven predictions, the framework facilitates informed decision-making in patient care.

\textbf{Strengths of the Study:}
One of the major strengths of this study is the robust performance of the QXAI framework in both vital sign prediction and physical activity classification tasks. The superior performance, as compared to traditional models, highlights the potential of deep learning in enhancing healthcare diagnostics and monitoring. Furthermore, the comprehensive nature of the explainability approach employed in this study marks a significant advancement over existing methods that typically focus on either post-hoc or intrinsic explainability. The practical application of the framework, demonstrated through its effectiveness on real-world datasets like PPG-DaLiA and MHEALTH, underscores its potential for implementation in real healthcare settings.

\textbf{Limitations and Future Directions:}
Despite its strengths, the study is not without limitations. The computational demands, particularly with large datasets due to the use of kernel SHAP, highlight the need for more efficient XAI algorithms. Additionally, while the framework shows promise, its generalizability across a broader range of healthcare scenarios remains to be tested. Future research should aim at scaling the framework for different types of healthcare data and conditions. Another area for future improvement is the user-centric design of the framework. Tailoring explanations to be intuitive for healthcare practitioners, with varying levels of technical expertise, could enhance its clinical adoption. Moreover, the integration of the framework within existing clinical workflows and ensuring data privacy and ethical AI use are crucial considerations for future development.

\section{Conclusion}\label{conclusion}
In healthcare applications, the explainability of machine learning model predictions or results is critical. This can assist clinicians in understanding the results to assist with clinical decisions that take appropriate steps for treatment. Existing deep learning models have a limitation in the explainability or interpretability of their results. The prediction or classification capacity of the proposed QXAI framework is outstanding compared to traditional machine learning models, with minimal knowledge of the healthcare domain knowledge to address the research problem. To utilize the advantage of the prediction capacity, this study proposed to adopt the Shapley values concept to vital signs prediction and decode global explanation at the overall population and local explanation at the subject level. However, the study was limited by the kernel SHAP method, which required significant memory and storage for large datasets. Future directions include incorporating more diverse feature inputs to enhance remote monitoring systems for clinical decision support.




\bibliographystyle{ieeetr}
\bibliography{ieee}

\begin{thebibliography}{10}

\bibitem{malasinghe2019remote}
L.~P. Malasinghe, N.~Ramzan, and K.~Dahal, ``Remote patient monitoring: a
  comprehensive study,'' {\em Journal of Ambient Intelligence and Humanized
  Computing}, vol.~10, no.~1, pp.~57--76, 2019.

\bibitem{asiimwe2020vital}
S.~B. Asiimwe, E.~Vittinghoff, and M.~Whooley, ``Vital signs data and
  probability of hospitalization, transfer to another facility, or emergency
  department death among adults presenting for medical illnesses to the
  emergency department at a~large urban hospital in the united states,'' {\em
  The Journal of Emergency Medicine}, vol.~58, pp.~570--580, apr 2020.

\bibitem{tao2022multi}
X.~Tao and J.~D. Velasquez, ``Multi-source information fusion for smart health
  with artificial intelligence,'' 2022.

\bibitem{prakash2020mapping}
N.~Prakash, A.~Manconi, and S.~Loew, ``Mapping landslides on eo data:
  Performance of deep learning models vs. traditional machine learning
  models,'' {\em Remote Sensing}, vol.~12, no.~3, p.~346, 2020.

\bibitem{muddamsetty2022visual}
S.~M. Muddamsetty, M.~N. Jahromi, A.~E. Ciontos, L.~M. Fenoy, and T.~B.
  Moeslund, ``Visual explanation of black-box model: similarity difference and
  uniqueness (sidu) method,'' {\em Pattern recognition}, vol.~127, p.~108604,
  2022.

\bibitem{Adadi2020}
A.~Adadi and M.~Berrada, ``Explainable {AI} for healthcare: From black box to
  interpretable models,'' in {\em Embedded Systems and Artificial
  Intelligence}, pp.~327--337, Springer Singapore, 2020.

\bibitem{lundberg2017unified}
S.~M. Lundberg and S.-I. Lee, ``A unified approach to interpreting model
  predictions,'' {\em Advances in neural information processing systems},
  vol.~30, 2017.

\bibitem{Gong2021}
K.~Gong, H.~K. Lee, K.~Yu, X.~Xie, and J.~Li, ``A prediction and interpretation
  framework of acute kidney injury in critical care,'' {\em Journal of
  Biomedical Informatics}, vol.~113, p.~103653, Jan. 2021.

\bibitem{Wu2018}
L.~Wu, Y.~Hu, X.~Liu, X.~Zhang, W.~Chen, A.~S.~L. Yu, J.~A. Kellum, L.~R.
  Waitman, and M.~Liu, ``Feature ranking in predictive models for
  hospital-acquired acute kidney injury,'' {\em Scientific Reports}, vol.~8,
  Nov. 2018.

\bibitem{ElShawi2020}
R.~ElShawi, Y.~Sherif, M.~Al-Mallah, and S.~Sakr, ``Interpretability in
  healthcare: A comparative study of local machine learning interpretability
  techniques,'' {\em Computational Intelligence}, vol.~37, pp.~1633--1650, Nov.
  2020.

\bibitem{ElShawi2019}
R.~Elshawi, M.~H. Al-Mallah, and S.~Sakr, ``On the interpretability of machine
  learning-based model for predicting hypertension,'' {\em {BMC} Medical
  Informatics and Decision Making}, vol.~19, July 2019.

\bibitem{ilic2021explainable}
I.~Ilic, B.~G{\"o}rg{\"u}l{\"u}, M.~Cevik, and M.~G. Baydo{\u{g}}an,
  ``Explainable boosted linear regression for time series forecasting,'' {\em
  Pattern Recognition}, vol.~120, p.~108144, 2021.

\bibitem{Bari2021}
D.~Bari{\'{c}}, P.~Fumi{\'{c}}, D.~Horvati{\'{c}}, and T.~Lipic, ``Benchmarking
  attention-based interpretability of deep learning in multivariate time series
  predictions,'' {\em Entropy}, vol.~23, p.~143, Jan. 2021.

\bibitem{Kaji2019}
D.~A. Kaji, J.~R. Zech, J.~S. Kim, S.~K. Cho, N.~S. Dangayach, A.~B. Costa, and
  E.~K. Oermann, ``An attention based deep learning model of clinical events in
  the intensive care unit,'' {\em {PLOS} {ONE}}, vol.~14, p.~e0211057, Feb.
  2019.

\bibitem{chen2023interpretable}
C.~Chen and B.~Li, ``An interpretable channelwise attention mechanism based on
  asymmetric and skewed gaussian distribution,'' {\em Pattern Recognition},
  p.~109467, 2023.

\bibitem{vaswani2017attention}
A.~Vaswani, N.~Shazeer, N.~Parmar, J.~Uszkoreit, L.~Jones, A.~N. Gomez,
  {\L}.~Kaiser, and I.~Polosukhin, ``Attention is all you need,'' {\em Advances
  in neural information processing systems}, vol.~30, 2017.

\bibitem{Bahdanau}
D.~Bahdanau, K.~Cho, and Y.~Bengio, ``Neural machine translation by jointly
  learning to align and translate,'' 2014.

\bibitem{dwivedi2023explainable}
R.~Dwivedi, D.~Dave, H.~Naik, S.~Singhal, R.~Omer, P.~Patel, B.~Qian, Z.~Wen,
  T.~Shah, G.~Morgan, {\em et~al.}, ``Explainable ai (xai): Core ideas,
  techniques, and solutions,'' {\em ACM Computing Surveys}, vol.~55, no.~9,
  pp.~1--33, 2023.

\bibitem{gunning2019darpa}
D.~Gunning and D.~Aha, ``Darpa’s explainable artificial intelligence (xai)
  program,'' {\em AI magazine}, vol.~40, no.~2, pp.~44--58, 2019.

\bibitem{zacharias2022designing}
J.~Zacharias, M.~von Zahn, J.~Chen, and O.~Hinz, ``Designing a feature
  selection method based on explainable artificial intelligence,'' {\em
  Electronic Markets}, pp.~1--26, 2022.

\bibitem{reiss2019deep}
A.~Reiss, I.~Indlekofer, P.~Schmidt, and K.~Van~Laerhoven, ``Deep ppg:
  large-scale heart rate estimation with convolutional neural networks,'' {\em
  Sensors}, vol.~19, no.~14, p.~3079, 2019.

\bibitem{Banos2014}
O.~Banos, R.~Garcia, J.~A. Holgado-Terriza, M.~Damas, H.~Pomares, I.~Rojas,
  A.~Saez, and C.~Villalonga, ``{mHealthDroid}: A novel framework for agile
  development of mobile health applications,'' in {\em Ambient Assisted Living
  and Daily Activities}, pp.~91--98, Springer International Publishing, 2014.

\bibitem{kingma2014adam}
D.~P. Kingma and J.~Ba, ``Adam: A method for stochastic optimization,'' {\em
  arXiv preprint arXiv:1412.6980}, 2014.

\bibitem{ni2019modeling}
J.~Ni, L.~Muhlstein, and J.~McAuley, ``Modeling heart rate and activity data
  for personalized fitness recommendation,'' in {\em The World Wide Web
  Conference}, pp.~1343--1353, 2019.

\bibitem{zhu2022fitness}
Z.~Zhu, H.~Li, J.~Xiao, W.~Xu, and M.-C. Huang, ``A fitness training
  optimization system based on heart rate prediction under different
  activities,'' {\em Methods}, vol.~205, pp.~89--96, 2022.

\bibitem{Shaik2022}
T.~Shaik, X.~Tao, N.~Higgins, R.~Gururajan, Y.~Li, X.~Zhou, and U.~R. Acharya,
  ``{FedStack}: Personalized activity monitoring using stacked federated
  learning,'' {\em Knowledge-Based Systems}, vol.~257, p.~109929, Dec. 2022.

\bibitem{Bozkurt2021}
F.~Bozkurt, ``A comparative study on classifying human activities using
  classical machine and deep learning methods,'' {\em Arabian Journal for
  Science and Engineering}, vol.~47, pp.~1507--1521, July 2021.

\bibitem{yijing2022prediction}
L.~Yijing, Y.~Wenyu, Y.~Kang, Z.~Shengyu, H.~Xianliang, J.~Xingliang, W.~Cheng,
  S.~Zehui, and L.~Mengxing, ``Prediction of cardiac arrest in critically ill
  patients based on bedside vital signs monitoring,'' {\em Computer Methods and
  Programs in Biomedicine}, vol.~214, p.~106568, 2022.

\bibitem{Ali2021}
M.~M. Ali, B.~K. Paul, K.~Ahmed, F.~M. Bui, J.~M. Quinn, and M.~A. Moni,
  ``Heart disease prediction using supervised machine learning algorithms:
  Performance analysis and comparison,'' {\em Computers in Biology and
  Medicine}, vol.~136, p.~104672, Sept. 2021.

\bibitem{malakar2022computer}
S.~Malakar, S.~D. Roy, S.~Das, S.~Sen, J.~D. Vel{\'a}squez, and R.~Sarkar,
  ``Computer based diagnosis of some chronic diseases: A medical journey of the
  last two decades,'' {\em Archives of Computational Methods in Engineering},
  pp.~1--43, 2022.

\bibitem{herm2023stop}
L.-V. Herm, K.~Heinrich, J.~Wanner, and C.~Janiesch, ``Stop ordering machine
  learning algorithms by their explainability! a user-centered investigation of
  performance and explainability,'' {\em International Journal of Information
  Management}, vol.~69, p.~102538, 2023.

\end{thebibliography}

\end{document}